\documentclass[10pt,twocolumn,letterpaper]{article}

\usepackage{iccv}
\usepackage{times}
\usepackage{epsfig}
\usepackage{graphicx}
\usepackage{amsmath}
\usepackage{amssymb}
\usepackage{multirow}
\usepackage{colortbl}
\usepackage{color}
\usepackage{threeparttable}
\usepackage{booktabs}
\usepackage{adjustbox}
\usepackage{subfig}
\usepackage{caption}

\usepackage[pagebackref=true,breaklinks=true,letterpaper=true,colorlinks,bookmarks=false]{hyperref}

 \iccvfinalcopy 

\def\iccvPaperID{1794} 

\ificcvfinal\fi

\newcommand{\RR}{\mathbb{R}}

\begin{document}

\title{Enhanced Boundary Learning for Glass-like Object Segmentation}

\author{
Hao He$^{1,2}$\thanks{Equal Contribution. Corresponding to: Lubin Weng, Guangliang Cheng} ,
Xiangtai Li$^{3*}$,
Guangliang Cheng$^{4,6}$,
Jianping Shi$^4$, \\
Yunhai Tong$^3$,
Gaofeng Meng$^{1,2,5}$,
V\'eronique Prinet$^1$,
Lubin Weng$^1$,
\\[0.2cm]
\small $ ^1$ National Laboratory of Pattern Recognition, Institute of Automation, Chinese Academy of Sciences \\
\small $ ^2$ School of Artificial Intelligence, University of Chinese Academy of Sciences, \small $ ^3$ Key Laboratory of Machine Perception (MOE), Peking University \\
\small $ ^4$ SenseTime Group Research
\small $ ^5$ Centre for Artificial Intelligence and Robotics, HK Institute of Science \& Innovation, CAS
\small $ ^6$ Shanghai AI Lab \\
\small\texttt{Email: hehao2019@ia.ac.cn, lxtpku@pku.edu.cn}
}

\maketitle
\ificcvfinal\thispagestyle{empty}\fi


\begin{abstract}
Glass-like objects such as windows, bottles, and mirrors exist widely in the real world. Sensing these objects has many applications, including robot navigation and grasping.
However, this task is very challenging due to the arbitrary scenes behind glass-like objects. This paper aims to solve the glass-like object segmentation problem via enhanced boundary learning. In particular, we first propose a novel refined differential module that outputs finer boundary cues. We then introduce an edge-aware point-based graph convolution network module to model the global shape along the boundary. We use these two modules to design a decoder that generates accurate and clean segmentation results, especially on the object contours. Both modules are lightweight and effective: they can be embedded into various segmentation models. In extensive experiments on three recent glass-like object segmentation datasets, including Trans10k, MSD, and GDD, our approach establishes new state-of-the-art results. We also illustrate the strong generalization properties of our method on three generic segmentation datasets, including Cityscapes, BDD, and COCO Stuff. Code and models is available at \url{https://github.com/hehao13/EBLNet}.
\end{abstract}
\section{Introduction}


\begin{figure}[!t]
	\centering
\includegraphics[scale=0.739]{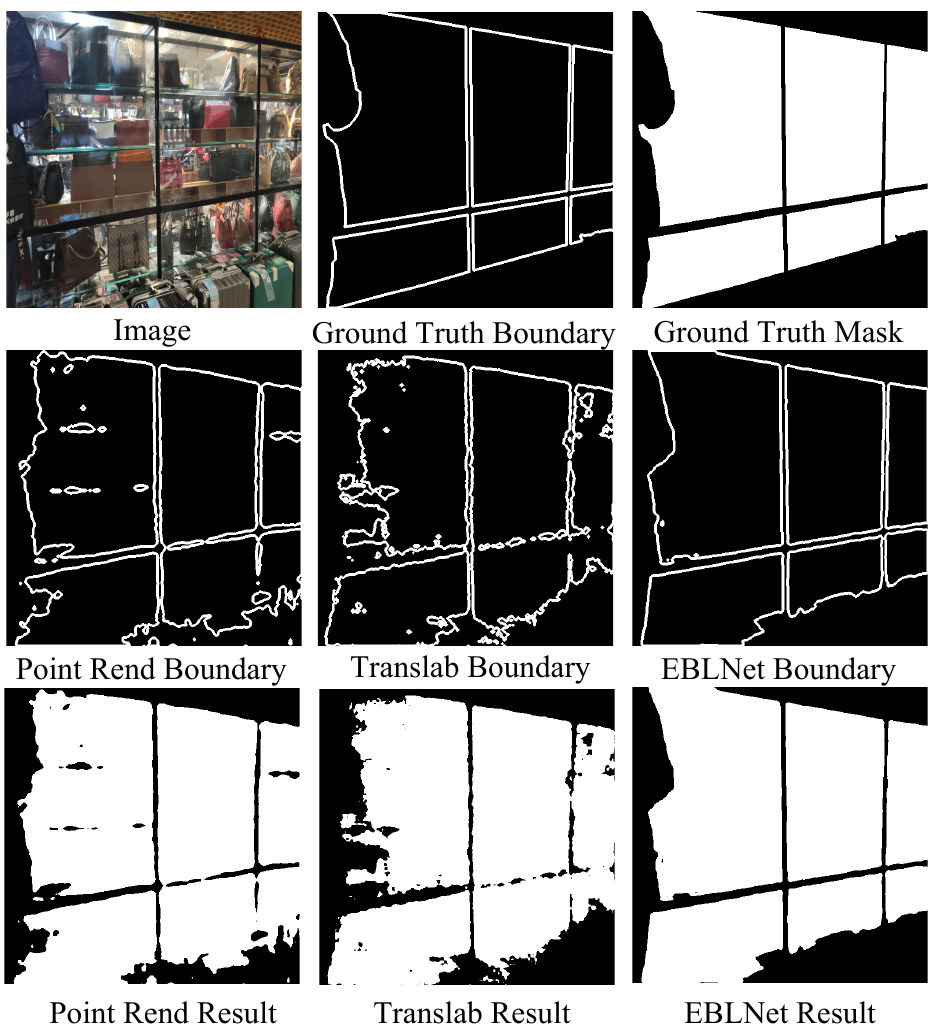}
\vspace{-2mm}
	\caption{
	\small Illustration of glass-like object segmentation. Row one gives the input image and ground truth. Row two shows the comparison results on the boundary. Row three compares the final segmentation results. Our method (EBLNet) works the best among them~\cite{kirillov2019pointrend,trans10k_xieenze}.
	}
	\label{fig:teaser}
	\vspace{-6mm}
\end{figure}
Glass-like objects widely exist in the real world: bottles, windows, and mirrors are made of glass. They are mostly transparent, often amorphous solid. Sensing these objects is useful for many applications. For example, robots need to avoid fragile objects (\textit{e.g.}, glasses, vases, and mirrors) during their navigation. These objects differ from others in that: (1) they do not have a fixed  pattern, their shape often vary, (2) because of their transparent nature, they are mostly confounded with the surrounding background, the appearance of their inner region  varying according to the scene, (3) not all glass regions are salient, sometimes there are some occlusions or reflections on it. These problems make such glass-like object segmentation a difficult task. Current state-of-the-art semantic segmentation models~\cite{pspnet,deeplabv3p,DAnet} and salient object detection methods~\cite{Poolnet,Bi_directional_message_pass} are significantly challenged. Similarly, though previous works on boundary detection~\cite{xie2015holistically,yu2017casenet,deng2018learning} achieve remarkable results on traditional benchmarks~\cite{silberman2012indoor,arbelaez2010contour}, they can not work well on glass-like object scenarios due to the confounding appearance of the object's inner and outer parts. Fig.~\ref{fig:teaser} illustrates a complex example for glass segmentation.

To address the above issues, the very first step is to construct large-scale datasets~\cite{tranparent_gdnet,Mirror_net,trans10k_xieenze}. From these datasets, the task is cast as a two-class semantic segmentation problem, where glass-like objects are foreground  and the rest is background.  Based on these newly proposed datasets, state-of-the-art methods of both semantic segmentation~\cite{DAnet,pspnet,deeplabv3p} and salient object detection~\cite{Poolnet,CPD} obtain unsatisfactory results. For example, DeeplabV3+~(semantic segmentation method)~\cite{deeplabv3p} and EGNet~(salient object detection method)~\cite{zhao2019egnet} only get 84.2 and 85.0 mIoU on GDD~\cite{tranparent_gdnet} test set. Very few dedicated methods to solve this challenging task have been introduced so far~\cite{tranparent_gdnet,trans10k_xieenze,Mirror_net}. GDNet~\cite{tranparent_gdnet} proposes to use multiple well-designed large-field contextual feature integration to enhance the context representation for modeling glass object context. MirrorNet~\cite{Mirror_net} uses a contextual contrasted feature extraction module for mirror detection. In contrast, Translab~\cite{trans10k_xieenze} proposes to use boundary-aware information to improve the segmentation performance. However, the first two~\cite{Mirror_net,tranparent_gdnet} bring too much comparison and context information, which is not very helpful. The last one ignores the pixel-wised relationship on  the boundary, limiting the generality of learning objects with various shapes.

When humans locate and recognize the glass-like object, one crucial cue is the entire boundary of such object. Instead, comparing the inner object appearance and its surrounding background used in MirrorNet and GDNet cannot help much to segment since they are mainly confounded. 

To reduce the influence of complex inner parts and get accurate boundary prediction, motivated by differential edge detection and morphological processing~\cite{evans2006morphological,papari2011edge}, we propose a novel and efficient module named Refined Differential Module (RDM).
RDM works in a coarse-to-fine manner with residual learning. Specifically, rather than simply predict the edge of objects with edge supervision, we also supervise the non-edge part following the same spirit of dilation operation in morphological processing~\cite{rivest1993morphological} to eliminate noise brought by the inner part and background. This process helps generate smoother non-edge features. Then the thinner and more accurate edge features can be generated by subtracting the non-edge features from the whole features. In order to use the precise edge prediction to enhance global feature learning around the edge then improve the final prediction, we further propose an efficient edge-aware Point-based Graph convolution network Module (PGM). In network design, our modules can be plugged into many semantic segmentation methods. Take DeeplabV3+~\cite{deeplabv3p} as an example, our modules can be inserted behind the ASPP module. Finally, we propose a joint loss function to simultaneously supervise the edge part, non-edge part, and final segmentation result.
 
{\em Quantitatively}, our approach achieves a significant gain in performance \wrt  previous work~\cite{kirillov2019pointrend,trans10k_xieenze} (around 3\%-5\% in mIoU), on three recent proposed glass-like object segmentation datasets  (Trans10k~\cite{trans10k_xieenze}, GDD~\cite{tranparent_gdnet}, and MSD~\cite{Mirror_net}), and hence achieves new state-of-the-art results. {\em Visually}, our method detects more accurate boundaries, resulting in a more precise segmentation result (see Fig.~\ref{fig:teaser}). This backs our original motivation that finer boundary prediction leads to more refined segmentation. 

We further verify the generalization performance of our model in Sec.~\ref{generalization experiments} via two different setting, one of them carries out the experiments on another three general segmentation datasets, including CityScapes~\cite{Cityscapes}, BDD~\cite{yu2020bdd100k}, and COCO Stuff~\cite{coco_stuff}. \\
Our main contributions are three-fold:
\begin{itemize}
	\item We analyze some existing methods, like DANet~\cite{DAnet}, Trans10k~\cite{trans10k_xieenze}, and GDNet~\cite{tranparent_gdnet} then identify their drawbacks for glass-like object segmentation task.
	\vspace{-2mm}
	\item We propose a Refined Differential Module~(RDM) to generate precise edge and an efficient Point-based Graph convolution network Module~(PGM) for global edge feature learning. Then a joint loss function is proposed to supervise the whole model.
	\vspace{-2mm}
	\item Extensive experiments and analyses indicate the effectiveness and generalization of our model. We achieve the state-of-the-art results on three challenging recent proposed glass-like object segmentation benchmarks. 
\end{itemize}
\section{Related Work}

\noindent
\textbf{Glass-like Object Segmentation:} Glass object segmentation comprises two sub-tasks, sometimes processed in a distinct way: mirror segmentation and transparent object segmentation. Giver a single RGB image, mirror segmentation aims to segment mirror regions. In~\cite{Mirror_net}, the author proposes a large benchmark for this topic. They attempt to use contextual contrasted features in a segmentation scheme. However, contrasted features may not help for transparent object segmentation since semantic and low-level discontinuities in mirrors may not happen for transparent objects. Transparent object segmentation aims to segment transparent objects from a single RGB image. In ~\cite{Kalra_2020_CVPR_transparent}, the author uses a polarization camera to capture multi-modal imagery to enhance the transparent object segmentation.
Authors of both ~\cite{tranparent_gdnet} and ~\cite{trans10k_xieenze} propose a new benchmark for this task. ~\cite{trans10k_xieenze} proposes Translab to encode the boundary information while ~\cite{tranparent_gdnet} integrates abundant contextual cues. Both of them have failed to explore the relationship between edge features and non-edge features, which is well-considered in our method.

\noindent
\textbf{Boundary Processing:} Boundary processing is a classical computer vision problem. In the era of deep learning, some CNN-based methods have significantly pushed the development of this field, such as~\cite{xie2015holistically, maninis2017convolutional, kokkinos2015pushing, deng2018learning}. Several works have been proposed to obtain satisfactory boundary localization with structure modeling, such as boundary neural fields~\cite{boundaries_network_fields}, affinity field~\cite{aaf} and random walk~\cite{cnn_random_wark}.
Recently, PointRend~\cite{kirillov2019pointrend} is proposed to refine the coarse masks by rending some points on them through one shared multi-layer perception module. Some works~\cite{Boundary_perserving_mask,xiangtl_decouple,he2021boundarysqueeze} fuse the learned edge map into segmentation head to improve segmentation performance on the boundary, while GSCNN~\cite{gated-scnn} uses a gated layer to control the information flow between edge part and regular part.  All these works focus on locally mining the edge information and ignore the global shape information, which is essential for recognizing the glass-like objects. Our PGM uses global edge features to improve the segmentation results of glass-like objects.

\noindent
\textbf{Semantic Segmentation:} 
Recent methods for semantic segmentation are predominantly based on FCNs~\cite{fcn,upernet,li2020semantic,xiangtl_gff}. Several works~\cite{deep_structured_seg,crf_as_rnn} use some structured prediction modules like conditional random fields~\cite{CRF} (dense CRFs) to refine outputs. Current state-of-the-art methods~\cite{pspnet,deeplabv3,DAnet} improve the segmentation performance by designing specific heads with dilated convolutions~\cite{dilation} to capture contextual information. Recently, non-local operators~\cite{non_local,DAnet,ccnet,zhang2019dual} based on the self-attention mechanism~\cite{vaswani2017attention} are used to harvest pixel-wise context from the whole image. Similarly, graph convolution networks~\cite{beyond_grids} propagate information over the entire image by reasoning in the interaction space.
 Although these methods achieve state-of-the-art results on standard benchmarks such as Cityscape~\cite{Cityscapes}, ADE~\cite{ADE20K}, BDD~\cite{yu2020bdd100k}, and COCO Stuff~\cite{coco_stuff}, their performance on glass-like object segmentation benchmarks will drop a lot due to noise of background context that are introduced by the non-local operators. Our method focuses more on the edges and avoids introducing too much background information.

\noindent
\textbf{Salient Object Detection:} Glass-like object segmentation can be viewed as one specific case of salient object detection. Many state-of-the-art methods~\cite{Bi_directional_message_pass,2016DHSNet, zhao2019egnet, pang2020multi, zhao2020suppress, gao2020highly} are devoted to fully utilize the integration of multi-level features to enhance network performance. Recent works~\cite{2017PiCANet, Chen_2018_ECCV} apply attention mechanisms to learn both global and local context or adopt foreground/background attention maps to help detect salient objects and eliminate non-salient objects. Our work also utilizes the multi-scale feature representation by designing a cascade differential refined model for better location cues to get fine-grained masks.
\begin{figure}[!t]
	\centering
	\includegraphics[scale=0.416]{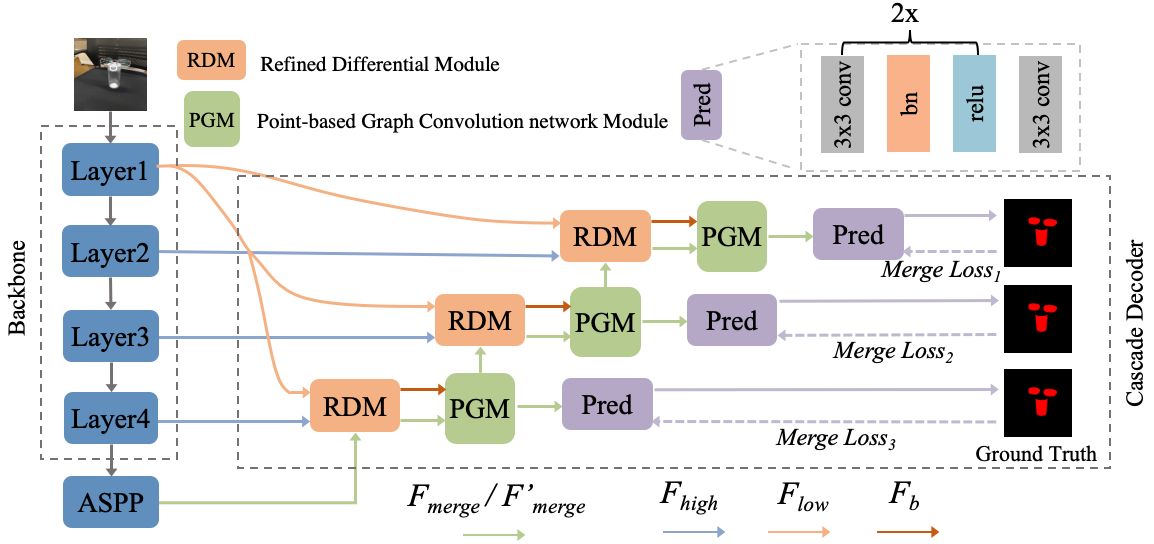}
	\caption{
	\small Cascade Network Architecture. We adopt DeeplabV3+ as baseline for illustration where the RDM and PGM are connected in a cascaded order. Best view it on screen and zoom in.
	}
	\label{fig:network}
	\vspace{-5mm}
\end{figure}

\section{Method}

\subsection{Overview}
\label{overview}

Our framework utilizes and combines the different levels of feature maps (high-level for semantic,  low-level for edge-like features) obtained from a pre-trained backbone network. Our observation has two-fold: (1)  edge information (from low-level feature maps) are necessary to `reinforce' semantic region cues (obtained from high-level maps), (2) precise and robust delineation of boundaries enables a fine and accurate segmentation result. These observations lead us to design two modules: a Refined Differential Module (\textbf{RDM}) and an edge-aware Point-based Graph convolution network Model~(\textbf{PGM}). RDM~(Sec.~\ref{RDM}) outputs the precise edges of the glass-like objects and the initial segmentation features. PGM~(Sec.~\ref{GCN}) learns to refine the initial segmentation features using the precise boundary generated by the RDM, it appears in coupling with the RDM. In Sec.~\ref{net}, RDM and PGM are integrated into a cascaded  architecture network.  The whole model is supervised in three fashions: boundaries-wise, residual-wise (nonboundary-wise), and merge-wise~(segmentation-wise) via a joint loss function. Fig.~\ref{fig:network} illustrates the overall framework. 

Note that, though Translab~\cite{trans10k_xieenze} also uses edge to improve the segmentation results, it has two main drawbacks: (1) directly predict the edge cannot avoid noises from glass objects' inner part, leading to undiscriminating edge feature and inaccurate edge prediction, (2) the direct usage of inaccurate edge prediction to improve segmentation feature will lead to some false guidance. Our two modules work in complementary ways to deal with these problems.

\subsection{Refined Differential Module}
\label{RDM}
\begin{figure}[!t]
	\centering
	\includegraphics[width=0.88\linewidth]{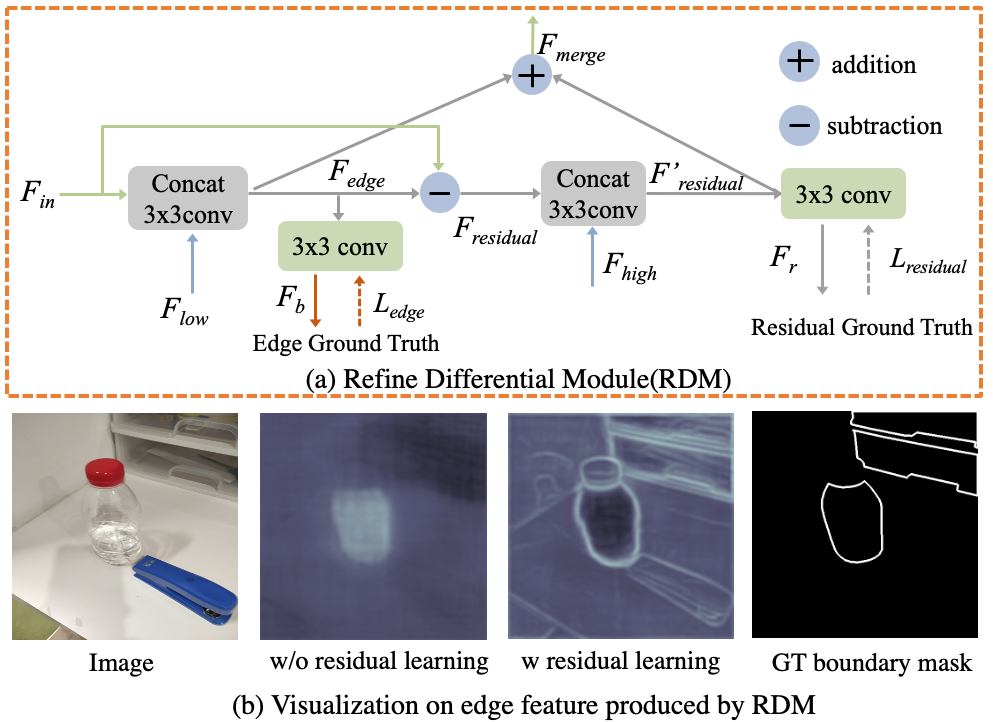}
	\caption{
	\small (a) Refined Differential Module, $L_{edge}$ and $L_{residual}$ are losses of edge part prediction and residual part prediction. (b) Feature visualization comparison. The second and the third column represent the edge feature $F_{edge}$ before and after using residual learning. Best view it on screen and zoom in.
	}
	\label{fig:RDE}
	\vspace{-5mm}
\end{figure}

\noindent \textbf{Overview:} The structure of RDM is shown in Fig.~\ref{fig:RDE}(a), inputs of the RDM are $F_{low}, F_{high}$ and $F_{in}$, which are low-level features of backbone, relative high-level features of backbone, and outputs of other modules~(such as ASPP module in DeeplabV3+~\cite{deeplabv3p}), respectively. Compared with $F_{high}$, $F_{in}$ has a larger receptive field and is smoother. With these three inputs, RDM aims to learn: (i) precise edge prediction, $F_{b}$, (ii) `residual' feature maps, $F_{residual}$ (segmentation features except the edge region), (iii) `complete' segmentation feature maps, $F_{merge}$, (that characterizes both residual region and edge region), which serves as the input of the next module. Supervision is applied for each feature type. Since there are two refinements and one subtraction~(which will be explained later) of our module, we name it Refined Differential Module or RDM for short.

\noindent \textbf{Coarse edge prediction:} This step  generates the edge prediction $F_b$ and non-edge features $F_{residual}$ . Since low-level features $F_{low}$ can provide detailed positional cues for edge prediction proven in previous works~\cite{deeplabv3p,kirillov2019pointrend}, we first \textbf{refine} the input feature $F_{in}$ with low-level features $F_{low}$ (after bi-linearly resizing the feature maps)  by first concat them. Then, a convolution operator is used to get the refined input feature, $F_{edge}\in \RR^{H \times W \times C}$, this process can be formulated by Equ.~\ref{equ:coarse_edge_predition}. 
\begin{equation}
   F_{edge}=g_{3\times3}([F_{low};F_{in}])
   \label{equ:coarse_edge_predition}
\end{equation}
where $[;]$ and $g_{n\times n}$ denote concatenation and convolution, respectively. Additionally, we compute  the residual (\ie  non-edge) feature $F_{residual}$ by \textbf{subtracting} $F_{edge}$ from $F_{in}$, \ie the difference between the coarse edge map $F_{edge}$ and the input features $F_{in}$. This procedure can be formulated as Equ.~\ref{equ:subtraction}: 
\begin{equation}
    F_{residual}=F_{in} - F_{edge},
    \label{equ:subtraction} 
\end{equation}
We further transform  $F_{edge} $ to the edge prediction $F_b  \in \RR^{H \times W \times 1}$ (via a 2-layer convolution), which is supervised using a ground-truth edge map. 

\noindent \textbf{Fine-grained Edge Prediction with Residual Learning:}
Our motivation has two-fold: (1) For the glass-like objects, the residual part may bring noises for edge prediction. Better feature representation of the residual part will make it easier to learn a better edge around the glass object. (2) Since $F_{edge}$ of the first step may not be very accurate, Equ.~\ref{equ:subtraction} may remove some vital features for $F_{residual}$. So we adopt another fine-grained feature $F_{high}$ generated from the backbone to \textbf{refine} $F_{residual}$ and get $F'_{residual}$ for better residual learning. 
This process is formulated by Equ.~\ref{equ:second refine}. 
\begin{equation}
   F'_{residual} = g_{3\times3}([F_{residual}; F_{high}])
   \label{equ:second refine}
\end{equation}
The prediction $F_r$ of the residual part is generated using another two $3 \times 3$ convolution layers, as shown in Equ.~\ref{equ:residual prediction}
\begin{equation}
	F_{r} = g_{3_\times3} \circ g_{3\times3}(F'_{residual})
	\label{equ:residual prediction}
\end{equation}
$F_{r}$ is supervised using a ground truth residual mask. This indirect supervision of $F'_{residual}$ forces it to be smooth, thus contributing to accurate $F_{b}$. The convolution operation here performs a similar role as the dilation operation in morphological processing~\cite{papari2011edge} (\ie making inner part `grow up' towards borders). 
Finally, we have $F_{merge}=F'_{residual} + F_{edge}$.
If there is no PGM connected to RDM, $F_{merge}$ is used to generate the prediction $F_m$ using a segmentation head consisting of three convolutional layers, two batch normalization layers~\cite{batchnorm}, and two non-linear activation function, shown in Fig.~\ref{fig:network}~(the purple module: Pred).

To best show the effect of our module, we give the feature visualization on $F_{edge}$ for both coarse edge prediction and fine-grained edge prediction with residual learning in Fig.~\ref{fig:RDE}(b). We use Principal component analysis~(PCA)~\cite{wold1987principal} to reduce the dimension of $F_{edge}$ into three dimensions for visualization. Compared to only supervising the edge part, joint supervision on both edge and residual part can produce more discriminative edge features, which supports our motivation for edge sharpening via residual learning.
 
 
\subsection{Edge-aware Point-based GCN Module}
\label{GCN}

\begin{figure}[!t]
	\centering
	\includegraphics[width=0.88\linewidth]{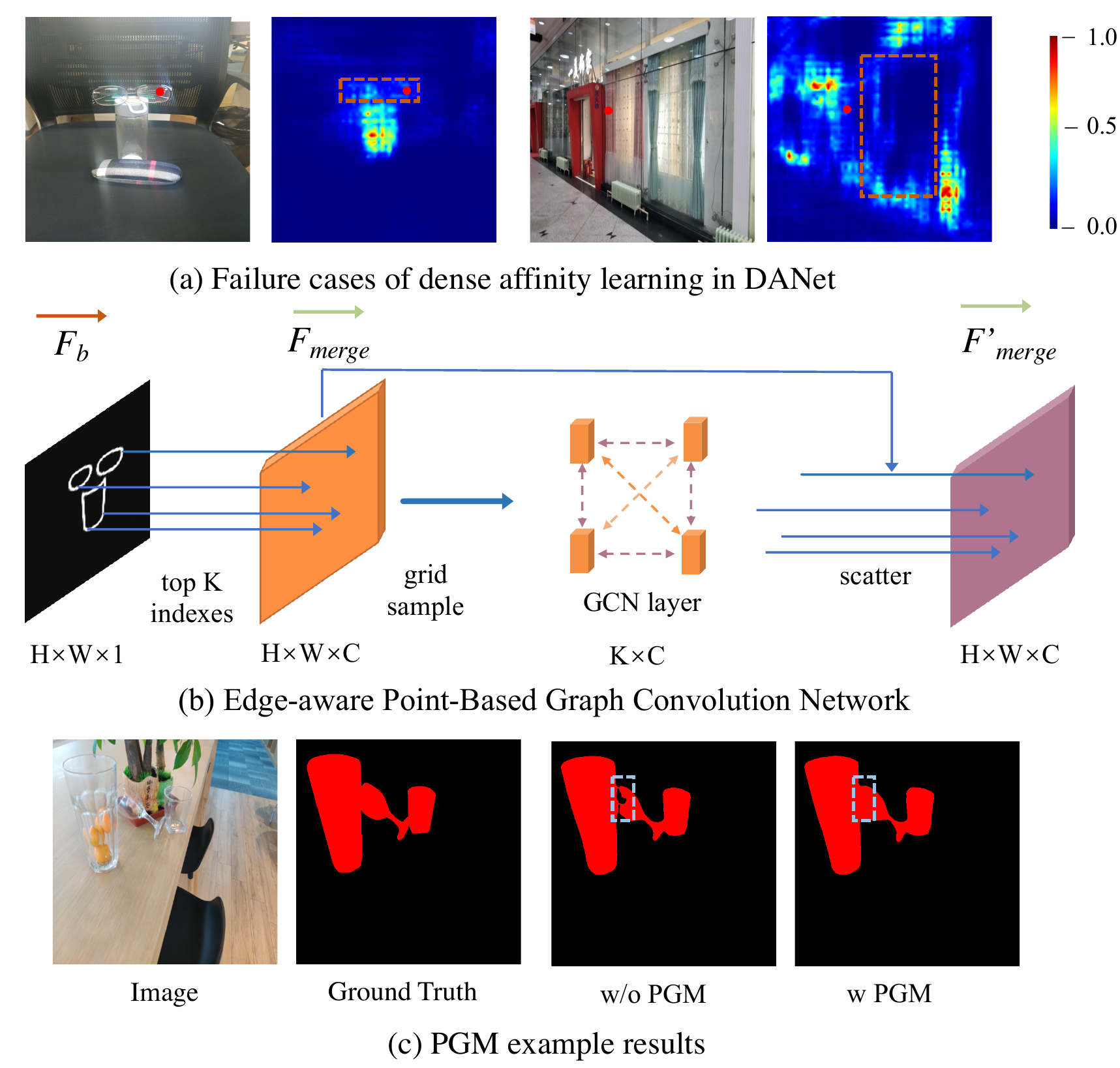}
	\vspace{-3mm}
	\caption{
	\small Subfigure (a) shows two failure cases for dense affinity learning in DANet~\cite{DAnet}, selected points are marked red. The second and the fourth images show the affinity map of selected points. For the first image, the red point is on the glass, while its affinity map mainly comes from the bottle not the glass. For the third image, the affinity map of the red point does not come from the glass to which it belongs. (b) describes the pipeline of PGM. (c) proves the effectiveness of PGM. After using PGM, the boundary is much smoother and more consistent. Best view it on screen.
	}
	\label{fig:pointGCN}
	\vspace{-5mm}
\end{figure}

In order to enforce finer boundary delineation on  $F_{merge}$, we further propose an efficient Point-based Graph convolution network~(GCN) Module. The key idea is to exploit the accurate edge prediction $F_b$ to refine the feature map $F_{merge}$ at the boundary points. To do so, our module accounts for the spatial correlation between edge points. We coin this module Point-based GCN Module (PGM).

\noindent
\textbf{Recall on Graph Convolution Network:} Graph convolution~\cite{gcnpaper} is a highly efficient and differentiable module that allows long-range interaction in a single operation.
Given a graph $\mathbb{G}=\{\mathbb{E}, \mathbb{V}\}$ where are $\mathbb{V}$ the nodes and  $\mathbb{E}$  the pairwise links between nodes, the graph convolution operation can be  defined as~\cite{gcnpaper} with Equ.~\ref{eq:gcn} :
\begin{equation}\label{eq:gcn}    
\tilde{\mathbf{X}} =  \sigma(\mathbf{W} \mathbf{X} \mathbf{A})
\end{equation}
with $\sigma(\cdot)$ a non-linear activation function, $\mathbf{A}\in \mathbb{R}^{N\times N}$ is the adjacency matrix characterising the neighbourhood relations of the graph, $\mathbf{W}\in\mathbb{R}^{D \times \tilde{D}}$ is a weight matrix and $\mathbf X$  is a $D \times N$ matrices characterizing the $N$ nodes in a $D$ dimensional space. However, directly applying GCN as dense affinity graph~\cite{DAnet,EMAnet} for glass-like object segmentation task is a trivial solution since the huge appearance variation on the surface of glass objects. As shown in Fig.~\ref{fig:pointGCN}(a), with the dense affinity graph, inner parts can not provide the context cues to support the final prediction on the visualized affinity maps. Thus we apply GCN only on point-wised boundary features of $F_{merge}$ to enhance its representation.

\noindent
\textbf{Point-based Graph Convolution Network on edge:} The pipeline of PGM is shown in Fig.~\ref{fig:pointGCN}(b). Given edge prediction $F_b$ and segmentation feature $F_{merge}$, we  sample the  nodes~$V$ from $F_b$ according to their confidence score and give them attributes to $F_{merge}$. Specifically, we first select at most $K$ points with the highest confidence in $F_{b}$. Then, we use the indices of these  points to sample the corresponding $K$ point features from $F_{merge}$. 
This allows us to build the feature matrix $\mathbf{G}_{in} \in \mathbb{R}^{C \times K} $, where $C$ is the number of channels of $F_{merge}$.  Assuming a fully connected graph (all $K$ points are  connected), the graph convolution operation can be performed as a fully connected layer. Then the adjacency matrix  $\mathbf{A}_g$  is simply a $1 \times 1$ convolutional layer that can be learned in a standard back-propagated way. Then we can write the output features with Equ.~\ref{eq:gcn1} :
\begin{equation}
\mathbf{G}_{out} = \sigma \left( \mathbf{W}_g \mathbf{G}_{in}  \mathbf{A}_g \right)
\label{eq:gcn1}
\end{equation}
with $\mathbf{A}_g \in \mathbb{R}^{K \times K }$.   $\mathbf{W}_g \in \mathbb{R}^{ C \times C }$ is a feature projection matrice learned along with  $\mathbf{A}$. 
After applying one layer of graph convolution and obtaining $\mathbf{G}_{out}$, we place back the point features of $G_{out}$ into $F_{merge}$ according to their indices and get the refined segmentation features $F'_{merge}$, which is used to generate the $F_{m}$ using the aforementioned Pred.

PGM is efficient and effective in handling glass-like transparent objects since the global edge information is essential to recognize a glass-like object. As shown in Fig.~\ref{fig:pointGCN}(c), after PGM, the final segmentation boundary is smoother and more consistent.

\subsection{Network Architecture and Loss function}
\label{net}


\noindent \textbf{Cascade Refined Decoder:} Cascade refinement has been proven effective for various tasks~\cite{cascade_rcnn,refinenet}. We also design a cascade decoder. For each layer, there are three modules~(RDM, PGM, Pred) connected serially. The outputs of RDM are $F_{b}$ and $F_{merge}$. Then PDM utilizes $F_{b}$ to improve the boundary part's feature of $F_{merge}$ and gets $F'_{merge}$. Finally, $F'_{merge}$ is used to generate the segmentation prediction $F_{m}$ via Pred. If it is not the last stage, we take the output of PGM $F'_{merge}$ as the new input $F_{in}$ for the next stage's RDM. The $F_{high}$ is designed by introducing different stages' features of the backbone network. We use the $F_{m}$ of the last cascade layer as the final output. Since our proposed modules are lightweight, cascade learning does not lead to much computation cost during the inference while contributes to a significant gain on final results. 

\noindent \textbf{Network Architecture:} Fig.~\ref{fig:network} illustrates the whole network architecture, which bases on the state-of-the-art semantic segmentation model DeeplabV3+~\cite{deeplabv3p}. Here we utilize dilated ResNet as backbone~\cite{resnet,dilation} only for illustration purpose. The output stride of backbone is 8 in our experiments unless otherwise specified. The cascade decoder is inserted after the ASPP module~\cite{deeplabv3p}, the output feature of ASPP module is served as the input feature $F_{in}$ in the first stage. Our method can be easily added to other methods~\cite{fcn,pspnet,DAnet} which can be found in the experiment part. 

\noindent \textbf{Loss function Design:} There are three different prediction results at each cascade stage, namely $F_b$, $F_r$, $F_m$, which are predictions of edge part, residual part and merge part, respectively. We define a joint loss function to supervise them simultaneously shown in Equ.~\ref{equ:joint_loss}.
\begin{equation}
    \label{equ:joint_loss}
    \begin{aligned}
    L_{joint} = &\lambda_{1}L_{residual}(F_r, G_r) + \lambda_{2}L_{edge}(F_b, G_e) + \\ &\lambda_{3}L_{merge}(F_m, G_m)
    \end{aligned}
\end{equation}
 $G_m$ represents the original ground truth, the ground truth of edge part $G_e$ generated from $G_m$ is a binary edge mask with thickness 8. When generating the ground truth of residual part $G_r$ the corresponding position of the positive pixel of $G_e$ in $G_r$ is marked as ignore and the other pixels of $G_r$ are the same as $G_m$. 
$\lambda_1$, $\lambda_2$ and $\lambda_3$ are used to balance the importance among $L_{residual}, L_{edge}$, $L_{merge}$. In our experiments, $\lambda_1=1, \lambda_2=3, \lambda_3=1$. $L_{merge}$ and $L_{residual}$ are standard Cross-Entropy Loss. $L_{edge}$ is Dice Loss~\cite{DiceLoss}. The final loss function is an addition of each stage, formulated as $ L=\sum_{n=1}^{N}L_{joint}^{n} $, where $N$ represents the number of cascade layer. Besides, in the PGM, we only use the indices of the selected edge points then the sample and place back operations can be implemented easily by some deep learning framework~(like grid sample and scatter in PyTorch~\cite{pytorch}). Thus, the overall model is end-to-end trainable.
\section{Experiment}

\begin{table*}[t]
	\begin{minipage}[!t]{\linewidth}
	    \centering
		\begin{minipage}{.45\linewidth}
		\subfloat[Effect of each component.]{
		\resizebox{1.0\textwidth}{!}{%
		\centering
		\begin{tabular}{c c c c l l l l}
					\hline
					 Baseline & +RDM & +cascade & +PGM & mIoU $\uparrow$ & mBER $\downarrow$ & mAE $\downarrow$  \\  
					\hline
				    Deeplabv3+	& - & - & - & 85.4 & 6.73 & 0.075 \\
					& \checkmark & - & - & 89.2 & 4.63 & 0.054 \\
					&\checkmark &\checkmark & - &  90.1 & 4.15 & 0.049 \\
					&\checkmark &\checkmark &\checkmark & \bf{90.7} & \bf{3.97} & \bf{0.047} \\
					\hline
				\end{tabular}}}\hspace{3mm}
	    \end{minipage}
	    \hspace{3mm}
	    \begin{minipage}{.45\linewidth}
		\subfloat[Effect of loss function design.]{
		
		\resizebox{0.95\textwidth}{!}{%
		\centering
		\begin{tabular}{c c c c l l l l}
					\hline
					 $L_{merge}$ & $L_{edge-bce}$ & $L_{edge-dice}$ & $L_{residual}$ &  mIoU $\uparrow$& mBER $\downarrow$ & mAE $\downarrow$  \\  
					\hline
					\checkmark & - & - & - & 88.3 & 5.02  & 0.060 \\
					\checkmark & \checkmark & - & & 89.4 & 4.48 & 0.053 \\
					\hline
				    \checkmark &\checkmark & -  &\checkmark & 90.4 & 4.12 & 0.049\\
					\checkmark & - &\checkmark & \checkmark & \bf{90.7} & \bf{3.97} & \bf{0.047} \\
					\hline
				\end{tabular}	
				
				}}\hspace{3mm}
		\end{minipage}
		
	\end{minipage}
    \vspace{3pt}
	\begin{minipage}[!t]{\linewidth}
	   \centering
		\begin{minipage}{.275\linewidth}
		\subfloat[Effect of each component in RDM. \label{expr:ablation_norm}]{
			\footnotesize
			\resizebox{0.95\textwidth}{!}{%
			\begin{tabular}{l|c c c}
				\hline
				Settings  & mIoU $\uparrow$ & mBER $\downarrow$ & mAE $\downarrow$ \\
				\hline
			    All operators & \bf{89.2} & \bf{4.63} & \bf{0.054} \\
				\hline
				w/o $F_{low}$ & 88.8 & 4.92 & 0.059 \\
				\hline
				w/o $F_{high}$& 88.4 & 5.01 & 0.058 \\
				\hline
				w/o Both &  88.2  & 4.98 & 0.058 \\
				\hline
				w/o residual learning & 86.4 & 6.17 & 0.070 \\
				\hline
		\end{tabular}}}
		\end{minipage}
        \begin{minipage}{.35\linewidth}
		\subfloat[Effectiveness on boundary refinement.]{
			\footnotesize
			\resizebox{0.95\textwidth}{!}{%
			\begin{tabular}{l c c c c c c }
				\toprule[0.15em]
				Method    & mIoU $\uparrow$  & F1(12px) $\uparrow$ & F(9px) $\uparrow$ & F1(5px) $\uparrow$ & F1(3px) $\uparrow$ \\
				\toprule[0.15em]
        baseline  & 85.4 & 77.6 & 73.5 & 66.6 & 56.7 \\
         +RDM &  89.2 & 84.7 & 81.6 & 77.6 & 68.5 \\
         +RDM \& PGM & 89.5 & 85.8 & 83.1 & 79.2 & 73.3 \\
	\bottomrule[0.1em]
	\end{tabular}}}
	    \end{minipage}
	    \begin{minipage}{.25\linewidth}
		\subfloat[Application on Other Architectures. \label{expr:ablation_placement}]{
			\footnotesize
			\resizebox{0.85\textwidth}{!}{%
			\begin{tabular}{l|c c c}
				\hline
				Network  & mIoU $\uparrow$ & mBER $\downarrow$ & mAE$\downarrow$ \\
				\hline
				FCN  & 83.2 & 7.94 & 0.094 \\
				\hline
			    +our modules &  87.9 & 5.04 & 0.062 \\
				\hline
				PSPNet~\cite{pspnet} & 83.5 & 8.34 & 0.089 \\
				\hline
				+our modules & 88.6 & 4.93 &  0.058 \\
				\hline
				DANet~\cite{DAnet} & 84.1 & 7.03 & 0.084 \\
				\hline
				+our modules & 87.8 & 5.14 & 0.062\\
				\hline
		\end{tabular}}}
		\end{minipage}
	\end{minipage}
	\vspace{-3mm}
	\caption{\small \textbf{Ablation studies.} We first verify the effect of each module and loss function in (a) and (b) in terms of mIoU, mBER and mAE. Then we give a detailed component analysis in RDM in (c), and use boundary metrics to verify each module's effect in (d). Finally, verify the various architectures in (e) to show the generality of our proposed modules. All the results are reported on the validation set of Trans10k. Best view it on screen and zoom in.}\label{tab:ablations}
	\vspace{-3mm}
\end{table*}


\subsection{Datasets and evaluation metrics}

\noindent \textbf{Datasets:} \textbf{(a) Trans10k}~\cite{trans10k_xieenze} dataset contains 10,428 images with two categories of transparent objects, including things and stuff. For this dataset, 5000, 1000, and 4428 images are used for train, validation, and test, respectively. 
\textbf{(b) GDD}~\cite{tranparent_gdnet} is another large-scale glass object segmentation dataset covering diverse daily-life scenes~(such as office, street). For dataset split, 2980 images are randomly selected for training and the remaining 936 images are used for testing.
\textbf{(c) MSD}~\cite{Mirror_net} is a large-scale mirror dataset. For dataset split, 3063 images are used for training while 955 images are used for testing.
We use the Trans10k for ablation studies and analyses and also report results on the GDD and MSD.

\noindent \textbf{Evaluation metrics:} Following the previous work~\cite{trans10k_xieenze, tranparent_gdnet, Mirror_net} , for the Trans10k dataset, we adopt four metrics for quantitatively evaluating the model performance unless otherwise specified. Specifically, we use the~(mean) intersection over union~(IoU/mIoU) and pixel accuracy~(ACC) metrics from the semantic segmentation field. Mean absolute error~(mAE) from the salient object detection field is also used. On top of that, we use the balance error rate~(BER/mBER), which considers the unbalanced areas of transparent~(mirror) and non-transparent~(non-mirror) regions. Following~\cite{tranparent_gdnet, Mirror_net}, F-measure is also used to evaluate the model performance of GDD and MSD datasets. When evaluating the accuracy of predicted boundaries, following~\cite{gated-scnn}, we use F1-score as the metric.

\subsection{Experiment on Trans10k}

\noindent \textbf{Overview:} We firstly perform ablation studies on the validation set. Then visualization analysis is given to show the accurate segmentation results. Finally, we compare our model with the state-of-the-art models on the test set.

\noindent \textbf{Trans10k implementation details:} We adopt the same training setting as the original Trans10k~\cite{trans10k_xieenze} paper for fair comparison. The input images are resized to the resolution of 512$\times$512. We use the pre-trained ResNet50 as the backbone and refine it along with other modules. We use 8 GPUs for all experiments and batch size is set to 4 per GPU. The learning rate is initialized to 0.01 and decayed by the poly strategy with the power of 0.9 for 16 epochs. We report mIoU, mBER, mAE in ablation study and also report ACC when comparing with other methods on test set.

\noindent \textbf{Ablation on Each Components:} We first verify the effectiveness of each module with DeeplabV3+~\cite{deeplabv3p} as the baseline in Tab.~\ref{tab:ablations}(a). After using the proposed RDM, there exists a significant gain over the baseline by 3.8\% mIoU, 2.1 in mBER and 0.021 in mAE. These results indicate the importance of precise boundaries generation for glass object segmentation. After appending more RDM modules in the cascade framework, the performance further gains 0.9\% mIoU. Finally, based on the precise boundary, applying our proposed PGM, our method gain another 0.6\% mIoU on the cascade model. Although PGM does not provide as much numerical improvement as RDM because of the small number of pixels at the boundaries, it does provide a significant visual improvement, as shown in Fig.~\ref{fig:pointGCN}(c). 

\noindent \textbf{Ablation on Loss function design:} Then we explore the impact of loss function by fixing each component where the number of sampled points is set to 96 in PGM, and the number of cascade stages is set to 3. Results are shown in Tab.~\ref{tab:ablations}(b), after using binary Cross-Entropy loss~($L_{edge-bce}$) as $L_{edge}$, we find 1.1\% mIoU gain. With an additional  $L_{residual}$, we find another 1.0\% mIoU gain which indicates that dual supervision is conducive to sharpen edge generation then improve the final prediction. Compared to using binary Cross-Entropy loss as $L_{edge}$, Dice loss~($L_{edge-dice}$) leads to 0.3\% mIoU gain since dice loss can better handle the imbalance problems for foreground objects. 


\noindent \textbf{Ablation on RDM:} We give detailed ablation studies on each component in RDM by removing it from the complete RDM in Tab.~\ref{tab:ablations}(c). We set the cascade number to 1 and do not use PGM. Removing $F_{low}$ or $F_{high}$ in the first step or the second step of RDM results in mIoU decline of 0.4\% and 0.8\%, respectively. Removing both of them results in mIoU degradation of 1.0\%. After removing the subtraction operation and residual learning, there is a clear drop on all metrics indicating the importance of boundary sharpening.

\noindent \textbf{Ablation on boundary refinement:} We present a detailed comparison on boundary using F1-score~\cite{gated-scnn} with four different thresholds in Tab.~\ref{tab:ablations}(d) where the cascade is not used. For each threshold, results obtain significant gains after using RDM. The gain becomes more significant when the neighbor pixels decrease (from 12 pixels to 3 pixels). These results indicate that our method can generate a precise boundary. The improvement of mIoU also proves our motivation that better boundary prediction is beneficial to produce better segmentation results. After using PGM, there are more gains in all the metrics in this strong baseline.

\noindent \textbf{Application on more segmentation methods:} In addition to DeeplabV3+, we carry on experiments on various other network architectures including,
 FCN~\cite{fcn}, PSPNet~\cite{pspnet} and DANet~\cite{DAnet}. Our modules are appended at the end of the head of these networks, and are trained with the same setting as DeeplabV3+. As shown in Tab.~\ref{tab:ablations}(e), our modules improve those methods by a large margin, which proves the generality and effectiveness of our modules.

\noindent \textbf{Comparison with the state-of-the-arts:} Following~\cite{trans10k_xieenze}, we report mIoU, Acc, mAE, mBER on Trans10k test set. As shown in Tab.~\ref{tab:results_trans10k}, our method achieves the state-of-the-art results on four metrics with output stride 16 for fair comparison with Translab~\cite{trans10k_xieenze}. After changing the output stride to 8 in the backbone, our method obtains a better result.


\begin{table}[!t]\setlength{\tabcolsep}{6pt}
	\centering
	\begin{threeparttable}
		\scalebox{0.63}{
			\begin{tabular}{l c  c c c  }
				\toprule[0.2em]
				Method   & mIoU $\uparrow$  &  Acc$\uparrow$  & mAE $\downarrow$  &  mBER$\downarrow$  \\
				\toprule[0.2em]
    Deeplabv3+~(MobileNetv2)~\cite{deeplabv3p} & 75.27 & 80.92 & 0.130 & 12.49 \\
    HRNet~\cite{HRNet} & 74.56 & 75.82 & 0.134 & 13.52 \\
    BiSeNet~\cite{bisenet} & 73.93 & 77.92 & 0.140 & 13.96 \\
    \hline
    DenseAspp~\cite{denseaspp} & 78.11 & 81.22 & 0.114 & 12.19 \\
    Deeplabv3+~(ResNet50)~\cite{deeplabv3p} & 84.54 & 89.54 & 0.081 & 7.78 \\
    FCN~\cite{fcn} & 79.67 & 83.79 & 0.108 & 10.33 \\
    RefineNet~\cite{refinenet} & 66.03 & 57.97 & 0.180 & 22.22 \\
    Deeplabv3+~(Xception65)~\cite{deeplabv3p} & 84.26 & 89.18 & 0.082 & 8.00 \\
    PSPNet~\cite{pspnet} & 82.38 & 86.25 & 0.093 & 9.72 \\
    Translab~(ResNet50)~\cite{trans10k_xieenze} & 87.63 & 92.69 & 0.063 & 5.46 \\
    \hline
    \hline
    EBLNet~(ResNet50, OS16)   & 89.58 & 93.95 & 0.052 & 4.60 \\
    EBLNet~(ResNet50, OS8)   & \bf{90.28} & \bf{94.71} & \bf{0.048} & \bf{4.14} \\
	\bottomrule[0.1em]
	\end{tabular}}
		\caption{\small Comparison to state-of-the-art on Trans10k test set. All methods are trained on Trans10k training set under the same setting and all models use single scale inference. OS means the output stride in the backbone. }
		\label{tab:results_trans10k}
	\end{threeparttable}
	\vspace{-3mm}
\end{table}

\subsection{Experiment on GDD and MSD}
\noindent \textbf{Experiment on GDD:} We adopt the same setting as the original paper~\cite{tranparent_gdnet} for fair comparison where the input images are resized into $416 \times 416$ and are augmented by horizontally random flipping. The total training epoch is 200. The learning rate is initialized to 0.003 and decayed by the poly strategy with the power of 0.9. We also give results with different backbones for better comparison. As shown in Tab.~\ref{tab:results_GDD}, our method achieves the state-of-the-art result on this dataset on five different metrics without using conditional random fields~(CRF)~\cite{CRF} as the post-processing.

\begin{table}[!t]\setlength{\tabcolsep}{6pt}
	\centering
	\begin{threeparttable}
		\scalebox{0.66}{
			\begin{tabular}{l c c c c c c }
				\toprule[0.2em]
				Method   & IoU $\uparrow$ &  Acc$\uparrow$  & $F_\beta \uparrow$ & mAE$\downarrow$  &  BER$\downarrow$  \\
				\toprule[0.2em]
	PSPNet~\cite{pspnet}  &  84.06 & 0.916 & 0.906  & 0.084  & 8.79 \\
    DenseASPP~\cite{denseaspp} & 83.68 &  0.919 & 0.911  & 0.081  & 8.66 \\
    DANet~\cite{DAnet} &84.15  & 0.911 & 0.901  &0.089  & 8.96  \\
    CCnet~\cite{ccnet} & 84.29 & 0.915 & 0.904 & 0.085 &  8.63\\
    PointRend~\cite{kirillov2020pointrend} & 86.51 & 0.933 & 0.928 & 0.067 & 6.50 \\
    \hline
    DSS~\cite{Hou_2017_CVPR} & 80.24 & 0.898 & 0.890 & 0.123 &  9.73\\
    PiCANet~\cite{2017PiCANet} & 83.73 & 0.916 & 0.909 &0.093  & 8.26 \\
    BASNet~\cite{qin2019basnet}& 82.88  & 0.907 & 0.896 & 0.094 &8.70  \\
    EGNet~\cite{zhao2019egnet}& 85.04 & 0.920  & 0.916 & 0.083 &  7.43 \\
    \hline
    DSC~\cite{Hu_2018_CVPR} & 83.56 & 0.914 & 0.911 & 0.090 & 7.97 \\
    BDRAR~\cite{Zhu_2018_ECCV} \textdagger  & 80.01  & 0.902 & 0.908 &  0.098 & 9.87 \\
    MirrorNet~\cite{Mirror_net}~(ResNext101)  \textdagger  & 85.07 & 0.918  & 0.903 & 0.083 &  7.67 \\
    GDNet~\cite{tranparent_gdnet}~(ResNext101) & 87.63  & 0.939 & 0.937 & 0.063 & 5.62 \\
    \hline
    \hline
    EBLNet~(ResNet101)   &  88.16 & 0.941 &  0.939 &  0.059  & 5.58 \\
    EBLNet~((ResNext101)   & \bf{88.72} & 	\bf{0.944} &  \bf{0.940} &	\bf{0.055} & \bf{5.36} \\
	\bottomrule[0.1em]
	\end{tabular}}
		\begin{tablenotes}
			 \item {\scriptsize \textdagger CRF is used for post-processing.
			 }
		\end{tablenotes}
		\caption{\small Comparison to state-of-the-art on GDD test set. All methods are trained on GDD training set under the same setting and all models use single scale inference. }
		\label{tab:results_GDD}
	\end{threeparttable}
\end{table}

\begin{table}[!t]\setlength{\tabcolsep}{6pt}
	\centering
	\begin{threeparttable}
		\scalebox{0.61}{
			\begin{tabular}{l c c c c c c c }
				\toprule[0.2em]
				Method  &  CRF & IoU $\uparrow$ &  Acc$\uparrow$  & $F_\beta \uparrow$ & mAE$\downarrow$  &  BER$\downarrow$  \\
				\toprule[0.2em]
	PSPNet~\cite{pspnet} & - & 63.21 & 0.750 & 0.746 & 0.117 & 15.82 \\
    ICNet~\cite{ICnet} & - & 57.25 & 0.694 & 0.710 & 0.124 & 18.75 \\
    PointRend~\cite{kirillov2020pointrend} (Deeplabv3+) & - & 78.81 & 0.936 & 0.872 & 0.054 & 8.95 \\
    \hline
    DSS~\cite{Hou_2017_CVPR} & - & 59.11 & 0.665 & 0.743 & 0.125 & 18.81 \\
    PiCANet~\cite{2017PiCANet} & - & 71.78 & 0.845 & 0.808 & 0.088 & 10.99 \\
    RAS~\cite{Chen_2018_ECCV} & - & 60.48 & 0.695 & 0.758 & 0.111 & 17.60 \\
    \hline
    DSC~\cite{Hu_2018_CVPR} & - & 69.71 & 0.816 & 0.812 & 0.087 & 11.77 \\
    BDRAR~\cite{Zhu_2018_ECCV} & $\surd$ & 67.43 & 0.821 & 0.792 & 0.093 & 12.41 \\
    MirrorNet~(ResNeXt101)~\cite{Mirror_net}  & $\surd$ & 78.95 & 0.935 & 0.857 & 0.065 & \bf{6.39} \\
    \hline
    \hline
    EBLNet~(ResNet101) & - & 78.84 & 0.946 & 0.873 & 0.054 & 8.84 \\
    EBLNet~(ResNext101) & - & \bf{80.33} & \bf{0.951} & \bf{0.883} & \bf{0.049} & 8.63 \\
	\bottomrule[0.1em]
	\end{tabular}
	}
	\caption{\small Comparison to state-of-the-art on MSD test set. All methods are trained on the MSD training set. All models use single-scale inference.}
	\label{tab:results_MSD}
	\end{threeparttable}
	\vspace{-2mm}
\end{table}

\begin{figure*}[!t]
	\centering
	\includegraphics[scale=0.439]{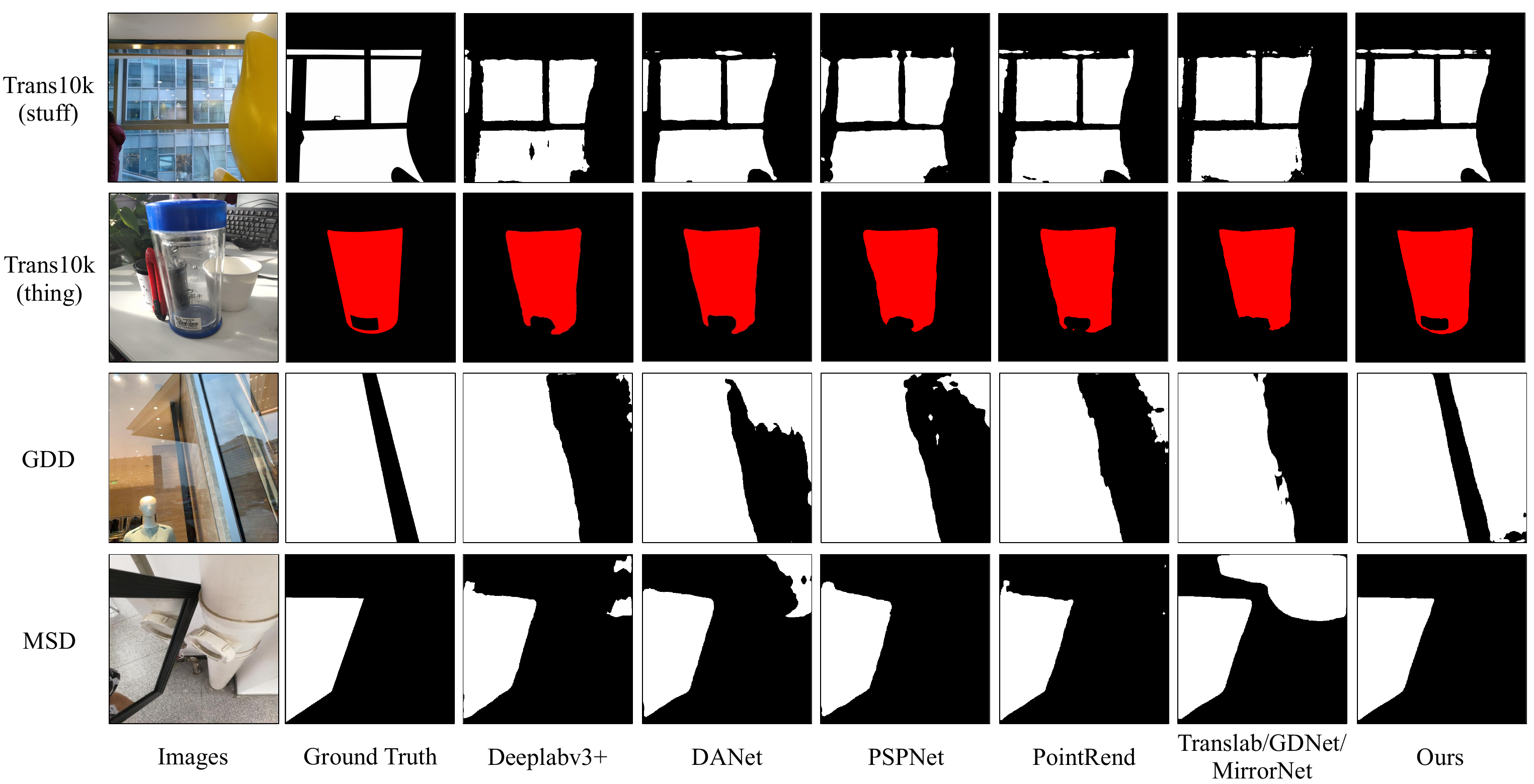}
	\vspace{-3mm}
	\caption{\small Visualization and comparison results on Three glass-like object segmentation datasets. The 1st and 2nd row of the 7th column are results of Translab~\cite{trans10k_xieenze}, the 3rd row of the 7th column is a result of GDNet~\cite{tranparent_gdnet}, the 4th row of the 7th column is a result of Mirrornet~\cite{Mirror_net}. Compared with other methods, our method has much better results.}
	\label{fig:trans10k_vis_res}
	\vspace{-2mm}
\end{figure*}

\noindent \textbf{Experiment on MSD:} Following the  setting in MirrorNet~\cite{Mirror_net}, input images are resized to a resolution of 384 $\times$ 384 and are augmented by horizontally random flipping. The learning rate is initialized to 0.002 and decayed by the poly strategy with the power of 0.9. We train our model for 160 epochs. As shown in Tab.~\ref{tab:results_MSD}, our method achieves state-of-the-art results in four different metrics.

\noindent \textbf{Visual Results Comparison:} Fig.~\ref{fig:trans10k_vis_res} gives the some visual results on all the three datasets. Our method can obtain superior segmentation results compared with other methods and has fine-grained boundary results on mirrors or glass. More results can be found in the supplemental file.

\begin{table}[!t]\setlength{\tabcolsep}{4pt}
	\centering
	\begin{threeparttable}
		\scalebox{0.66}{
		\begin{tabular}{c c c c l l l l}
					\hline
					 Method  & Parameters $\downarrow$ & Inference Time $\downarrow$ & mIoU $\uparrow$ & BER $\downarrow$ & mAE $\downarrow$   \\  
					\hline
					 MirrorNet~\cite{Mirror_net} & 103.3M & 37ms & 81.3 & 8.98 & 0.094 \\
				     GDNet~\cite{tranparent_gdnet} &183.2M & 41ms & 82.6 & 8.42 & 0.088 \\
				     Translab~\cite{trans10k_xieenze} & 65.2M & 26ms & 85.1 & 7.43 & 0.081 \\
					 EBLNet & \bf{46.2M} & \bf{24ms} & \bf{86.0} & \bf{6.90} & \bf{0.074} \\
					\hline
				\end{tabular}}
		\vspace{-2mm}
		\caption{\small Comparison results with related methods on GDD test set. The speeds are tested with $416 \times 416$ inputs. For fair comparison, all models' backbone are ResNet50 and all the inference times are tested with one V100 GPU.
		}
		\label{tab:comparison_results_with_efficienct}
	\end{threeparttable}
	\vspace{-5mm}
\end{table}

\noindent \textbf{Efficiency and Effectiveness:} To further show the efficiency and effectiveness of our model on glass-like object segmentation, we compare parameters and inference time of our network with recent proposed glass-like object segmentation networks, including MirrorNet~\cite{Mirror_net}, GDNet~\cite{tranparent_gdnet} and Translab~\cite{trans10k_xieenze} on GDD test set. Results are shown in Tab.~\ref{tab:comparison_results_with_efficienct}. Our method has the fewest parameters and fastest inference speed. Our method also achieves the best precision.

\subsection{Generalization Experiments}
\label{generalization experiments}
In this section, we prove the generality of our method via two different experiment settings: (1)~training our model on one glass-like object segmentation dataset then fine-tune on another glass-like object segmentation dataset, (2)~evaluating our model on other three general segmentation datasets, including Cityscapes~\cite{Cityscapes}, BDD~\cite{yu2020bdd100k} and COCO Stuff~\cite{coco_stuff}. Training settings can be found in the supplemental file. 
\begin{table}[!t]
	\begin{center}
	\scalebox{0.66}{
	\begin{tabular}{c c c c |c c c c}
		\hline
		OD & FTD & mIoU & SFT mIoU & OD & FTD & mIoU & SFT mIoU \\
		\hline
		Trans10k & MSD & 78.43 & 78.84 & MSD & Trans10k & 86.42 & 90.28 \\
		Trans10k & GDD & 87.94 & 88.16 & GDD & Trans10k & 89.01 & 90.28\\
		MSD & GDD & 87.07 & 88.16 & GDD & MSD & 79.41 & 78.84\\
		\hline
	\end{tabular}}
	\end{center}
	\vspace{-2mm}
		\caption{\small OD: original dataset. FTD: fine-tune dataset. Models are pre-trained on OD then fine-tuned on FTD and report mIoU on the test set of FTD. FTD mIoU: when model is trained from scratch on the FTD, the mIoU on the FTD test set. The mIoUs are very close to SFT mIoUs. Note that, the mIoU even \textbf{surpass} SFT mIoU when OD is GDD and FTD is MSD. Backbones are ResNet101.}
	\label{table:Generation evaluation1}
	\vspace{-2mm}
\end{table}
\begin{table}[!t]\setlength{\tabcolsep}{4pt}
	\centering
	\begin{threeparttable}
		\scalebox{0.6}{
			\begin{tabular}{l c c c c c c c c }
				\toprule[0.2em]
				Method   & dataset & mIoU $\uparrow$ & F1(12px) $\uparrow$ & F1(9px) $\uparrow$ & F1(5px) $\uparrow$ & F1(3px) $\uparrow$ \\
				\toprule[0.2em]
		Deeplabv3+ \cite{deeplabv3p} & Cityscapes & 77.4 & 78.9 & 77.5 & 73.9 & 62.3 \\
    	+PointRend \cite{pspnet}  &  Cityscapes & 78.3 & 79.5 &  78.5 & 74.3 & 63.6 \\
        +ours   &  Cityscapes & 79.1 & 81.4 & 80.1 & 76.5 & 67.0  \\
        \hline
        Deeplabv3+ \cite{deeplabv3p} & BDD & 60.8 & 76.4 & 75.2 & 70.6 & 61.5\\
    	+PointRend \cite{pspnet}  & BDD & 61.2 & 78.4 & 77.2 & 72.3 & 63.6 \\
        +ours   &  BDD  & 63.1 & 80.6 & 79.4 & 75.4 & 66.4  \\
        \hline
        Deeplabv3+ \cite{deeplabv3p} &  COCO Stuff & 33.6 & 73.3 & 71.8 & 70.6 & 66.7\\
    	+PointRend \cite{pspnet}  & COCO Stuff & 34.1 & 73.8 & 72.3 & 71.1 & 67.4 \\
        +ours   &  COCO Stuff  & 34.7 & 74.9 & 72.9 & 71.5 & 67.7 \\
	\bottomrule[0.1em]
	\end{tabular}}
	\vspace{-2mm}
		\caption{\small Comparison results on Cityscapes, BDD and COCO Stuff datasets with DeeplabV3+ and PointRend where X-px means X pixels along the boundaries. All the models are trained and tested with the same setting. Backbones are ResNet50.
		}
		\label{tab:results_on_cityscapes_bdd}
		\vspace{-3mm}
	\end{threeparttable}
\end{table}

\noindent \textbf{Fine-tune Setting:} In this setting, we only fine-tune the models for \textbf{few} epochs in the new datasets~(2, 10, 8 epochs for Trans10k, GDD and MSD, respectively). Results are shown in Tab.~\ref{table:Generation evaluation1}, which indicates our method is \textbf{not} overfitted to one specific glass-like object segmentation dataset.

\noindent \textbf{General Segmentation Dataset Setting:} In this setting, we use the DeeplabV3+ as the baseline method. Results are shown in Tab.~\ref{tab:results_on_cityscapes_bdd}. For all the three datasets, both mIoU and F1 scores of our method are better than those of DeeplabV3+ and PointRend. Thus, our method is \textbf{not} overfitted to glass-like object segmentation dataset.
\section{Conclusion}
In this paper, we focus on solving the glass-like object segmentation problem by enhancing the boundary learning for existing semantic segmentation methods. We propose a novel Refined Differential Module for edge prediction and an efficient edge-aware Point-wised Graph convolution network Module to model global shape representation of glass objects and guide the final prediction. We achieve the new state-of-the-art results on three recent glass-like object segmentation datasets, including Trans10k, MSD, and GDD. Further experiments on three general segmentation datasets, including Cityscapes, BDD, and COCO Stuff prove the generality and superiority of our method.
\section{Acknowledgement}
This research was supported by the National Key Research and Development Program of China under Grant No. 2018AAA0100400, No. 2020YFB2103402, and the National Natural Science Foundation of China under Grants 62071466, 62076242, and 61976208.

{\small
\bibliographystyle{ieee_fullname}
\bibliography{egbib}
}

\end{document}